\documentclass[11pt]{article}

\PassOptionsToPackage{sort&compress}{natbib}

\usepackage[preprint]{acl}

\usepackage{times}
\usepackage{latexsym}

\usepackage[T1]{fontenc}
\usepackage[utf8]{inputenc}

\usepackage{microtype}
\usepackage{nicefrac}

\usepackage{inconsolata}

\usepackage{graphicx}
\graphicspath{ {./figures/} }

\usepackage{mathtools} % for math macros

\DeclarePairedDelimiter{\abs}{\lvert}{\rvert}

\usepackage[noabbrev]{cleveref} % for clever references
\usepackage{subcaption} % for subfigures

\title{Emergence of Minimal Circuits for Indirect Object Identification in Attention-Only Transformers}

\author{Rabin Adhikari\\
Saarland University\\
66123 Saarbrücken \\
\href{mailto:raad00002@stud.uni-saarland.de}{raad00002@stud.uni-saarland.de} }

\begin{document}
    \maketitle

    \begin{abstract}
    Mechanistic interpretability aims to reverse-engineer large language models (LLMs) into human-understandable computational circuits. However, the complexity of pretrained models often obscures the minimal mechanisms required for specific reasoning tasks. In this work, we train small, attention-only transformers from scratch on a symbolic version of the Indirect Object Identification (IOI) task, a benchmark for studying coreference-like reasoning in transformers. Surprisingly, a single-layer model with only two attention heads achieves perfect IOI accuracy, despite lacking MLPs and normalization layers. Through residual stream decomposition, spectral analysis, and embedding interventions, we find that the two heads specialize into additive and contrastive subcircuits that jointly implement IOI resolution. Furthermore, we show that a two-layer, one-head model composes information from the previous layer primarily through query-key interactions. These results demonstrate that task-specific training induces highly interpretable, minimal circuits, offering a controlled testbed for probing the computational foundations of transformer reasoning.

\end{abstract}
    \section{Introduction}

Large Language Models (LLMs) have achieved remarkable success across a vast range of natural language processing tasks \citep{DBLP:journals/corr/abs-2601-03267, DBLP:journals/corr/abs-2503-19786, DBLP:journals/corr/abs-2512-02556, DBLP:journals/corr/abs-2407-21783, DBLP:journals/corr/abs-2505-09388, DBLP:journals/corr/abs-2601-08584}. Yet, their internal operations remain largely opaque, motivating the field of mechanistic interpretability, which seeks to reverse-engineer these ``black boxes'' into understandable circuits and algorithms \citep{olah2020zoom,DBLP:conf/iclr/NandaCLSS23,cammarata2020thread}. Its ultimate goal is to achieve a circuit-level understanding where individual components like neurons and attention heads are mapped to specific algorithmic roles~\citep{elhage2021mathematical,DBLP:conf/nips/ConmyMLHG23}. However, the immense scale, residual connections, and non-linearities of modern LLMs present significant challenges to this endeavor.

To navigate this complexity, researchers often start with simplified or ``toy'' models as controlled environments for developing and validating interpretability tools \citep{geva-etal-2021-transformer,DBLP:conf/iclr/NandaCLSS23,DBLP:journals/corr/abs-2209-10652,DBLP:conf/icml/ChughtaiCN23,heimersheim2023circuit,DBLP:journals/tmlr/FurutaMIM24}. By training models on constrained, synthetic objectives, we can reduce confounding variables from complex linguistic structures and discover core computational mechanisms in a cleaner setting.

A common approach to understanding these models involves analyzing pre-trained transformers~\citep{DBLP:conf/nips/VaswaniSPUJGKP17} on specific tasks they can perform \citep{brinkmann-etal-2024-mechanistic,DBLP:journals/tmlr/BereskaG24}. To investigate these capabilities, we focus on the Indirect Object Identification (IOI) task. \citet{DBLP:conf/iclr/WangVCSS23} showed that GPT-2 small~\citep{radford2019language} implements IOI through a multi-hop attention circuit involving distinct classes of heads. However, this mechanism arises within a model pretrained for next-token prediction on natural text, which is an inherently complex optimization setting.

In contrast, we train minimal, attention-only transformer models \citep{DBLP:conf/nips/VaswaniSPUJGKP17} from scratch exclusively on a symbolic version of the IOI task. We find that a straightforward model, a single-layer transformer block with just two attention heads, can solve this task perfectly. Because the IOI task requires dynamic duplicate-token detection and exclusionary copying, the computations proven to be beyond the representational capacity of bigram and skip-gram models \citep{elhage2021mathematical,DBLP:conf/iclr/WangVCSS23}, our findings build on \citet{buck2023one} by demonstrating exactly how a minimal model with a single attention layer implements this logic. Furthermore, a detailed analysis of this model uncovers a highly compact and interpretable circuit where the solution is computed via a direct additive combination of the two heads' outputs, rather than a complex, multi-hop pipeline found in GPT-2 small \citep{radford2019language}.

Our contributions are threefold:
\begin{enumerate}
    \item We demonstrate that a one-layer, two-head attention-only model is sufficient to solve the IOI task with a fixed template perfectly.

    \item We provide a mechanistic analysis that uncovers a minimal circuit based on an additive combination of specialized attention head outputs.

    \item We argue that the circuits in large, broadly pre-trained models may be overly complex due to multi-task pressures, whereas task-constrained training can reveal more parsimonious mechanisms.
\end{enumerate}
    \section{Background}

\subsection{Task: Indirect Object Identification (IOI)}
The IOI task serves as a standardized benchmark for studying coreference-like reasoning and dynamic memory mechanisms within language models \citep{DBLP:conf/iclr/WangVCSS23,ensign2024investigating}. In a typical natural language IOI sentence, an initial dependent clause introduces two distinct names: a subject (S) and an indirect object (IO). Subsequently, the main clause repeats the subject. The model's objective is to accurately predict the IO as the next logical token. For example, in the sentence ``When John and Mary went to the store, John gave a drink to \_\_\_,'' the model must predict ``Mary''. Following \citet{DBLP:conf/iclr/WangVCSS23}, which identified a complex, multi-hop circuit for this task in GPT-2 small~\citep{radford2019language}, our work investigates this fundamental exclusionary logic in a drastically simplified setting.

\subsection{Transformer Architecture}
To establish our mathematical notation, we rely on the framework introduced by \citet{elhage2021mathematical} for reverse-engineering attention-only transformers. This framework conceptualizes the transformer's residual stream as a primary communication channel where different components read and write information independently.

\subsubsection{Residual Stream Decomposition}

For an attention-only model, the state of the residual stream at layer $l$, denoted as $x^{(l)}$, is strictly a linear combination of the initial embeddings and the outputs of all preceding attention heads. Let $x_{embed}$ and $x_{pos}$ denote the token and positional embeddings, respectively. The residual stream just before unembedding in a single-layer model is formalized as follows.
\[
    x^{(1)}= x_{embed}+ x_{pos}+ \sum_{h=1}^{H}\text{out}_{h}^{(0)}
\]
where $\text{out}_{h}^{(0)}$ is the output vector written to the residual stream by head $h$ in layer $0$. The final logit prediction for any token $t$ is computed by projecting this residual stream onto the unembedding matrix $W_{U}$: $L_{t}= (x^{(1)})^{T}W_{U[:, t]}$.

\subsubsection{Transformer Circuits}

To compute its output, each attention head $h$ reads from the residual stream using three projection matrices: the Query matrix ($W_{Q}^{h}$), the Key matrix ($W_{K}^{h}$), and the Value matrix ($W_{V}^{h}$). The queries and keys interact to determine the attention scores between tokens, while the values determine the information moved across the sequence, which is then projected back into the residual stream via an output matrix ($W_{O}^{h}$).

We can analyze the behavior of individual heads by decomposing these operations into two distinct circuits \citep{elhage2021mathematical}. The $QK$ (Query-Key) circuit dictates the attention scores between a query token and a key token, represented by the end-to-end transition matrix $M_{QK}^{h}= W_{E}^{T}(W_{Q}^{h})^{T}W_{K}^{h}W_{E}$, where $W_{E}$ denotes the token embedding matrix. The $OV$ (Output-Value) circuit dictates how attending to a specific token updates the final output logits, represented by the transition matrix $M_{OV}^{h}= W_{U}W_{O}^{h}W_{V}^{h}W_{E}$. Because these exact matrices, $M_{QK}^{h}$ and $M_{OV}^{h}$, map directly from the vocabulary space back to the vocabulary space, they serve as the foundation for our spectral analysis in \cref{sec:spectral_analysis}.

\subsubsection{Composition of Attention Heads}
In multi-layer attention-only architectures, transformer heads develop functional hierarchies by composing information across layers \citep{elhage2021mathematical}. Because the input to a head in a subsequent layer $j$ is the residual stream $x^{(j)}$, which contains the outputs of heads from an earlier layer $i$ ($i < j$), the projection matrices of layer $j$ directly read the information written by layer $i$. This interaction is formally categorized into three types of composition described below.

\paragraph{Q-Composition}
The output of Layer $i$ is projected through the Query matrix of Layer $j$, affecting what the latter head attends to.

\paragraph{K-Composition}
The output of Layer $i$ is projected through the Key matrix of Layer $j$, altering how the latter head matches incoming queries.

\paragraph{V-Composition}
The output of Layer $i$ is projected through the Value matrix of Layer $j$. This modifies the actual information the later head moves across the sequence, effectively creating a ``virtual attention head.''

While both Q- and K-composition enable more complex attention routing by acting on different sides of the attention score calculation, V-composition strictly affects information transfer. We leverage this framework in \cref{sec:composition_ablation} to perform targeted ablations, identifying precisely which pathways our two-layer model relies upon to solve the IOI task.
    \section{Dataset and Model Configuration}

\subsection{The IOI Task in a Symbolic Setting}
To isolate the core relational reasoning challenge of the IOI task described above, we construct a purely symbolic dataset. This formulation abstracts away all linguistic and tokenization complexities, enabling precise inspection of what the model must represent to distinguish between ``subject'' and ``object'' roles without natural language confounds.

The training data consists of $6$-token sequences following the format \texttt{<BOS> IO S1 S2 <MID> ?}. The \texttt{IO}, \texttt{S1}, and \texttt{S2} are two unique names drawn from a small vocabulary. The model's task is to predict the name token that is not repeated before the \texttt{<MID>} token.

Essentially, the dataset follows two rigid templates, depending on the order of the names. Using ``John'' and ``Mary'' as examples, the templates are:
\begin{enumerate}
    \item \texttt{<BOS> John Mary Mary <MID> John}

    \item \texttt{<BOS> John Mary John <MID> Mary}
\end{enumerate}
Following \citet{DBLP:conf/iclr/WangVCSS23}, we refer to the first template as ``BABA'' and the second as ``BAAB''.

\subsection{Model Configuration}
\label{sec:model_configuration}

To maximize interpretability, we used simple attention-only transformer models with absolute positional embeddings. Feed-forward networks and layer normalization were omitted to isolate the function of the attention mechanism. The vocabulary consists of $6$ name tokens plus the two special tokens \texttt{<BOS>} and \texttt{<MID>}, for a total size of $8$. The residual stream dimension was kept the same size as the vocabulary ($8$), and the head dimension was $d_{head}= \nicefrac{8}{N_h}$, where $N_{h}$ is the number of heads. In this formulation, the number of parameters in the model is independent of the number of heads.

\subsubsection{Training and Analysis Setup}
Models were trained from scratch on the symbolic IOI dataset using a cross-entropy loss to predict the token at the \texttt{<MID>} position. Each training batch contained all $60$ possible unique sequences ( $6 \times 5 = 30$ ways of picking the names in the dependent clause and two ways of ordering them in the main clause). We used the AdamW~\citep{DBLP:conf/iclr/LoshchilovH19} optimizer with the OneCycle~\citep{smith2019super} learning rate scheduler, with a maximum learning rate of $0.1$ and weight decay of $0.01$. Training and analyses were performed on a single NVIDIA A40 GPU using PyTorch~\citep{DBLP:conf/nips/PaszkeGMLBCKLGA19} and TransformerLens~\citep{nanda2022transformerlens} libraries.
    \section{Results and Analysis}

\subsection{Zero-Layer and Single-Head Baselines}

\begin{figure}[h!]
    \begin{subfigure}
        {\linewidth}
        \includegraphics[width=\linewidth]{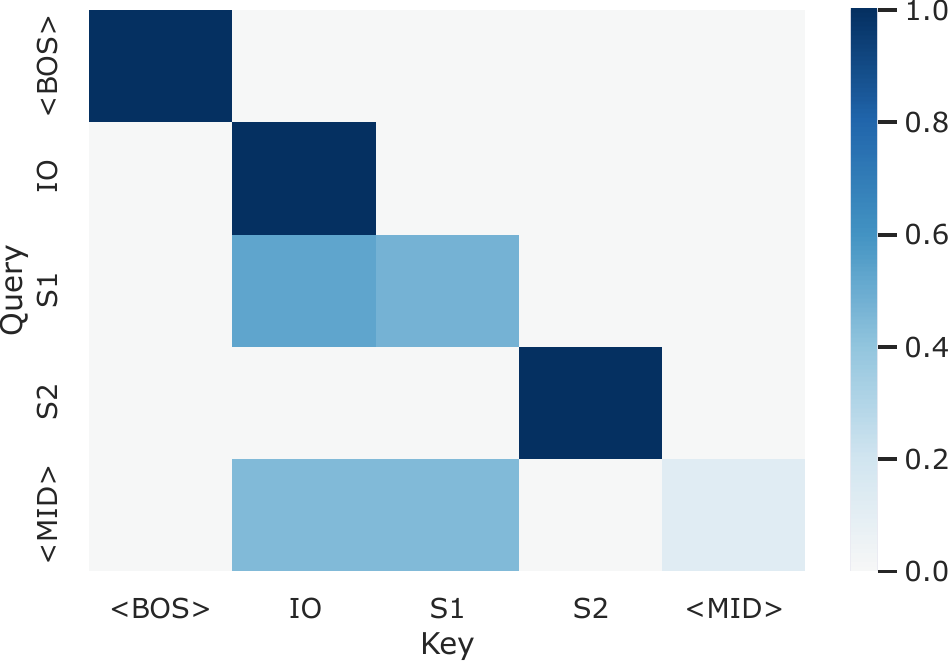}
        \caption{Average Attention Heatmap}
        \label{fig:attention_patterns_ioi_layers_1_heads_1}
    \end{subfigure}
    \begin{subfigure}
        {\linewidth}
        \includegraphics[width=\linewidth]{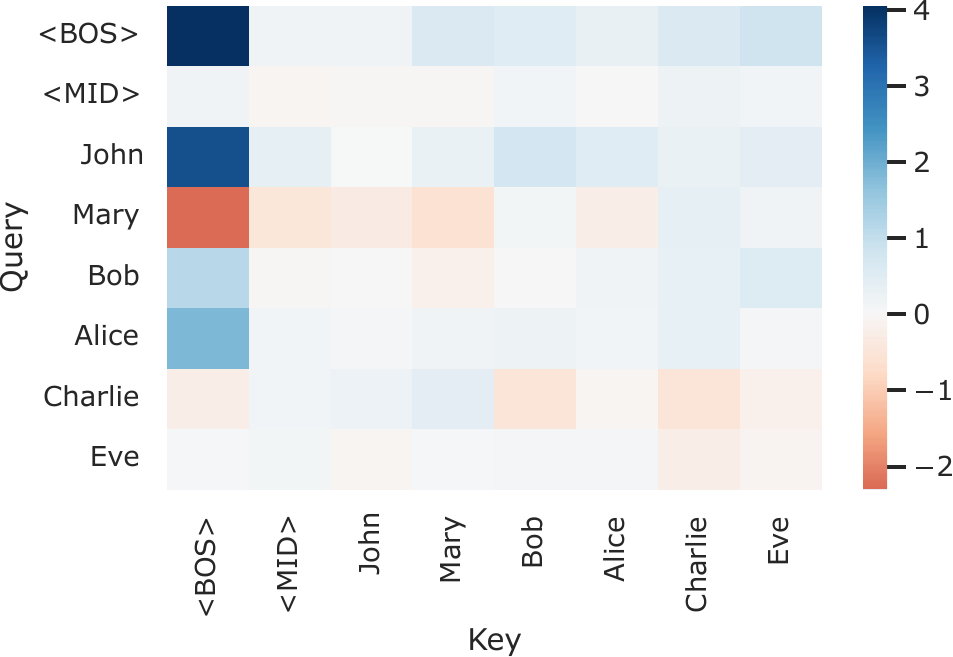}
        \caption{The QK Circuit}
        \label{fig:qk_circuit_layers_1_heads_1}
    \end{subfigure}
    \begin{subfigure}
        {\linewidth}
        \includegraphics[width=\linewidth]{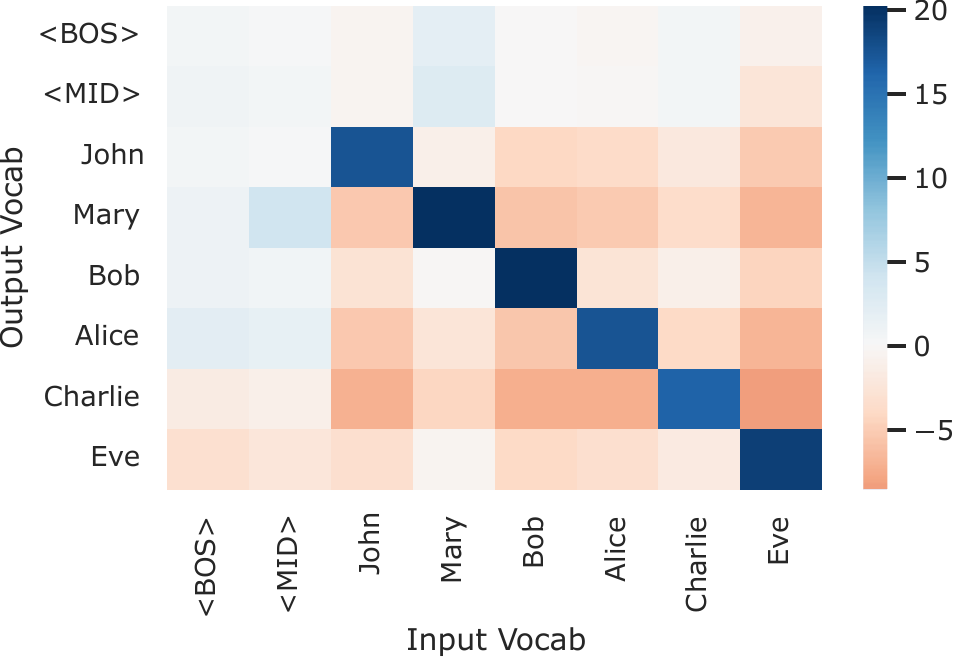}
        \caption{The OV Circuit}
        \label{fig:ov_circuit_layers_1_heads_1}
    \end{subfigure}
    \caption{Single-head, one-layer models fail at the IOI task. The attention and circuit analysis reveal that a single head cannot simultaneously detect the correct reference and copy it to the output.}
    \label{fig:single_head_one_layer_ioi}
\end{figure}

A zero-layer model (acting as a bigram) predicts every name token with an equal probability of $\approx \nicefrac{1}{N_{names}}= 16.7\%$, as the \texttt{<MID>} token must predict a name without utilizing prior context. When extending to a single-layer, single-head model, it assigns $\approx0.5$ probability for the names provided in the prompt. However, it cannot distinguish which one is correct. As shown in \cref{fig:attention_patterns_ioi_layers_1_heads_1}, the \texttt{<MID>} token attends roughly equally to both the names in the dependent clause, indicating that a single attention head cannot jointly encode the information required to \textbf{(i)} identify which token serves as the correct referent and \textbf{(ii)} propagate that information to the prediction position. The roles of ``reference detection'' and ``copying'' appear to be functionally incompatible within the attention mechanism with a single head.

We analyze the QK and OV circuits to understand the failure mode of the single-head model. From the QK circuit (see \cref{fig:qk_circuit_layers_1_heads_1}), we observe that the \texttt{<MID>} token attends almost uniformly to all tokens. And the OV circuit (see \cref{fig:ov_circuit_layers_1_heads_1}) shows that each name token makes a high positive contribution to its own logit but small negative contributions to the logits for other names. The uniform attention pattern averages these contributions, resulting in similar logits for both names.

\subsection{A Two-Heads, One-Layer Model Learns IOI Perfectly}

\begin{figure}[h]
    \begin{subfigure}
        {\linewidth}
        \includegraphics[width=\linewidth]{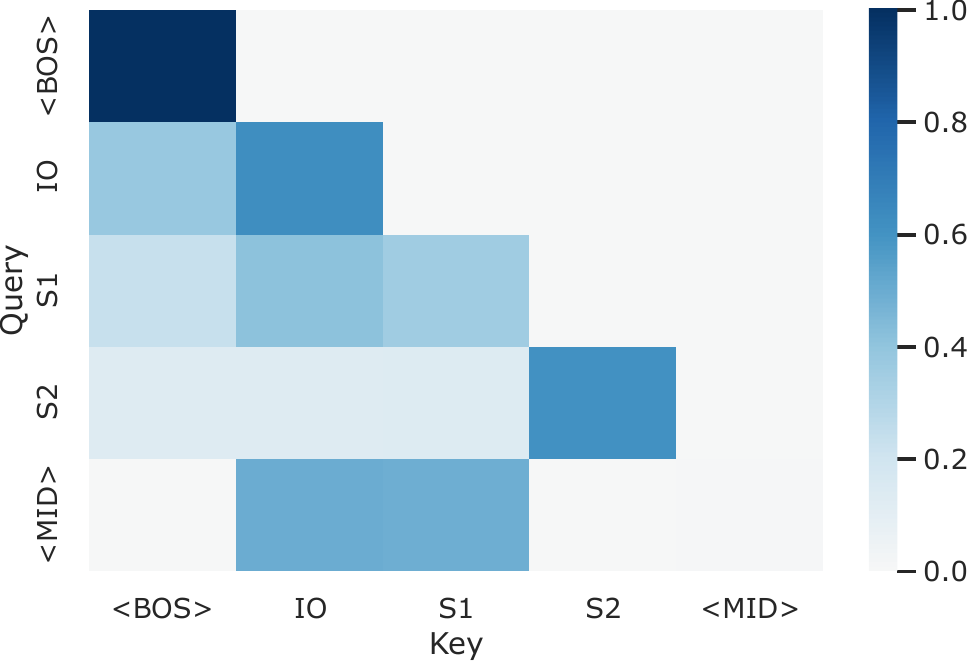}
        \caption{The first head attends equally to both context names.}
        \label{fig:attention_patterns_ioi_layers_1_heads_2_head0}
    \end{subfigure}
    \begin{subfigure}
        {\linewidth}
        \includegraphics[width=\linewidth]{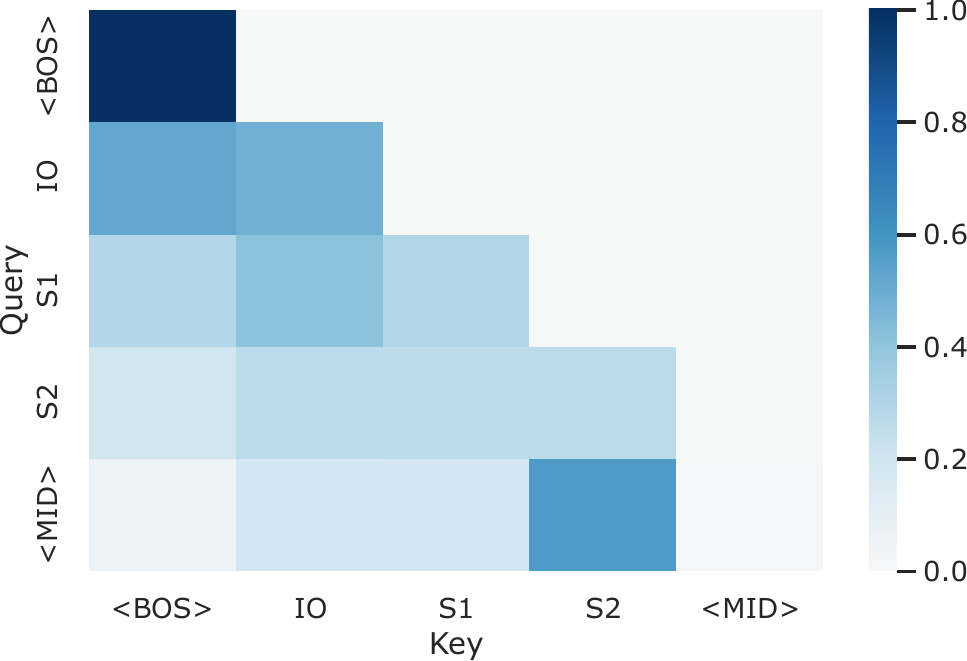}
        \caption{The second head distributes attention between the main clause subject and the names in the dependent clause.}
    \end{subfigure}
    \caption{Average attention heatmap for a two-heads, one-layer model.}
    \label{fig:attention_patterns_ioi_layers_1_heads_2}
\end{figure}

When the model with a single attention layer is extended to two attention heads, it achieves perfect accuracy on the IOI task. \Cref{fig:attention_patterns_ioi_layers_1_heads_2} shows distinct attention patterns of the two heads, averaged across all the possible inputs.

% \begin{figure}[h]
%     \centering
%     \includegraphics[width=\linewidth]{attention_patterns_ioi_layers_1_heads_2}
%     \caption{\textbf{Average Attention Heatmap for Two-Head, One-Layer Model.} The first head focuses almost equally on the two name tokens from the dependent clause, while the second head has half of its attention on the subject of the main clause and almost a quarter on each of the names in the dependent clause.}
%     \label{fig:attention_patterns_ioi_layers_1_heads_2}
% \end{figure}

\subsubsection{Two Heads with Distinct Roles}

\begin{figure}[h]
    \begin{subfigure}
        {\linewidth}
        \includegraphics[width=\linewidth]{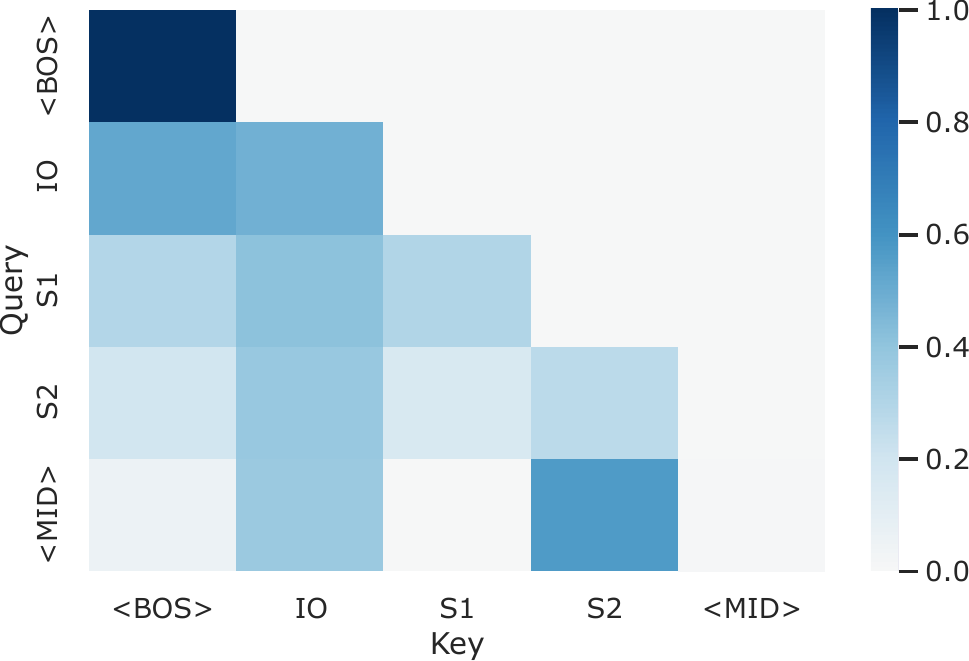}
        \caption{The second head attends to \texttt{IO} and \texttt{S2} for the ``BAAB'' template.}
    \end{subfigure}
    \begin{subfigure}
        {\linewidth}
        \includegraphics[width=\linewidth]{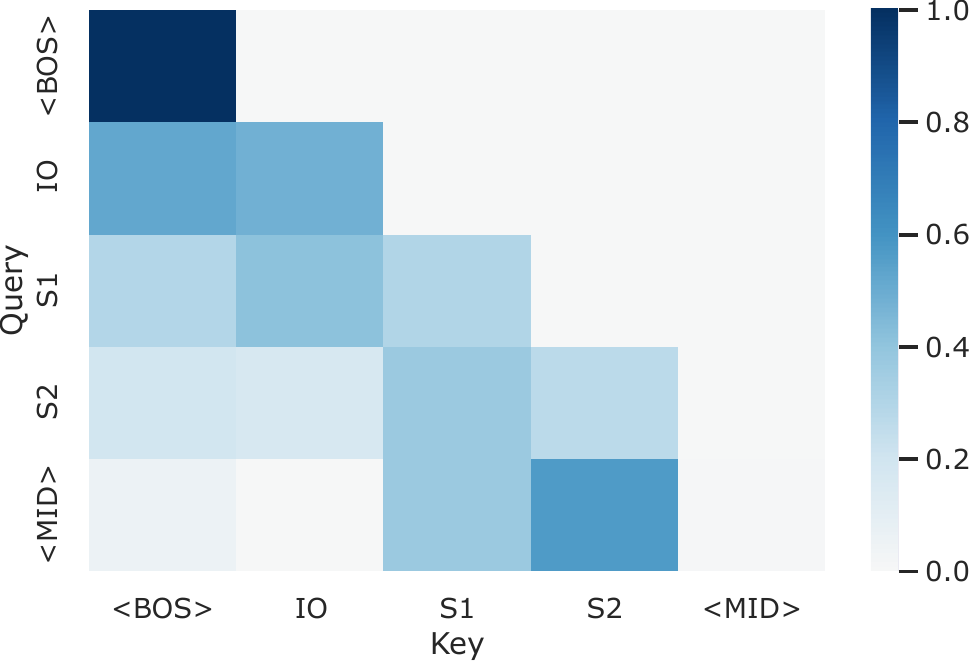}
        \caption{The second head attends to \texttt{S1} and \texttt{S2} for the ``BABA'' template}
    \end{subfigure}
    \caption{The attention map for the second head depends on the template. It dynamically attends to the subject of the main clause and the non-repeated name of the dependent clause.}
    \label{fig:attention_patterns_ioi_layers_1_heads_2_combined}
\end{figure}

We observed that, for both the templates of our symbolic dataset, the first head consistently attends to the two names in the initial dependent clause (see \cref{fig:attention_patterns_ioi_layers_1_heads_2}), indicating its role in identifying the relevant referents. The second head, however, always attends to the subject of the main clause and the non-repeated name in the dependent clause (see \cref{fig:attention_patterns_ioi_layers_1_heads_2_combined}). So, this head does most of the heavy lifting, finding out the unique set of tokens to attend to. Furthermore, it suggests that the second head is responsible for integrating the referential information with the context provided by the main clause to determine the correct output.

% \begin{figure}[h]
%     \centering
%     \includegraphics[width=\linewidth]{attention_patterns_ioi_layers_1_heads_2_combined}
%     \caption{\textbf{The attention map for the second head depends on the template.} While the first head always attends to the two name tokens in the dependent clause, the second head attends to the second occurrence of the subject in the main clause and the other name in the dependent clause --- ``BA'' in the ``BAAB'' template and similarly, ``AB'' in the ``BABA'' template.}
%     \label{fig:attention_patterns_ioi_layers_1_heads_2_combined}
% \end{figure}

% \begin{figure}[h]
%     \includegraphics[width=\textwidth]{attention_patterns_baba_layers_1_heads_2}
%     \caption{\textbf{Average Attention Heatmap for Two-Head, One-Layer Model on BABA Template.} The first head focuses almost equally on the two name tokens from the dependent clause, while the second head attends to the IO as well as the subject of the main clause.}
%     \label{fig:attention_patterns_baba_layers_1_heads_2}
% \end{figure}

\subsubsection{Residual Stream Decomposition}

\begin{figure}[h]
    \centering
    \includegraphics[width=\linewidth]{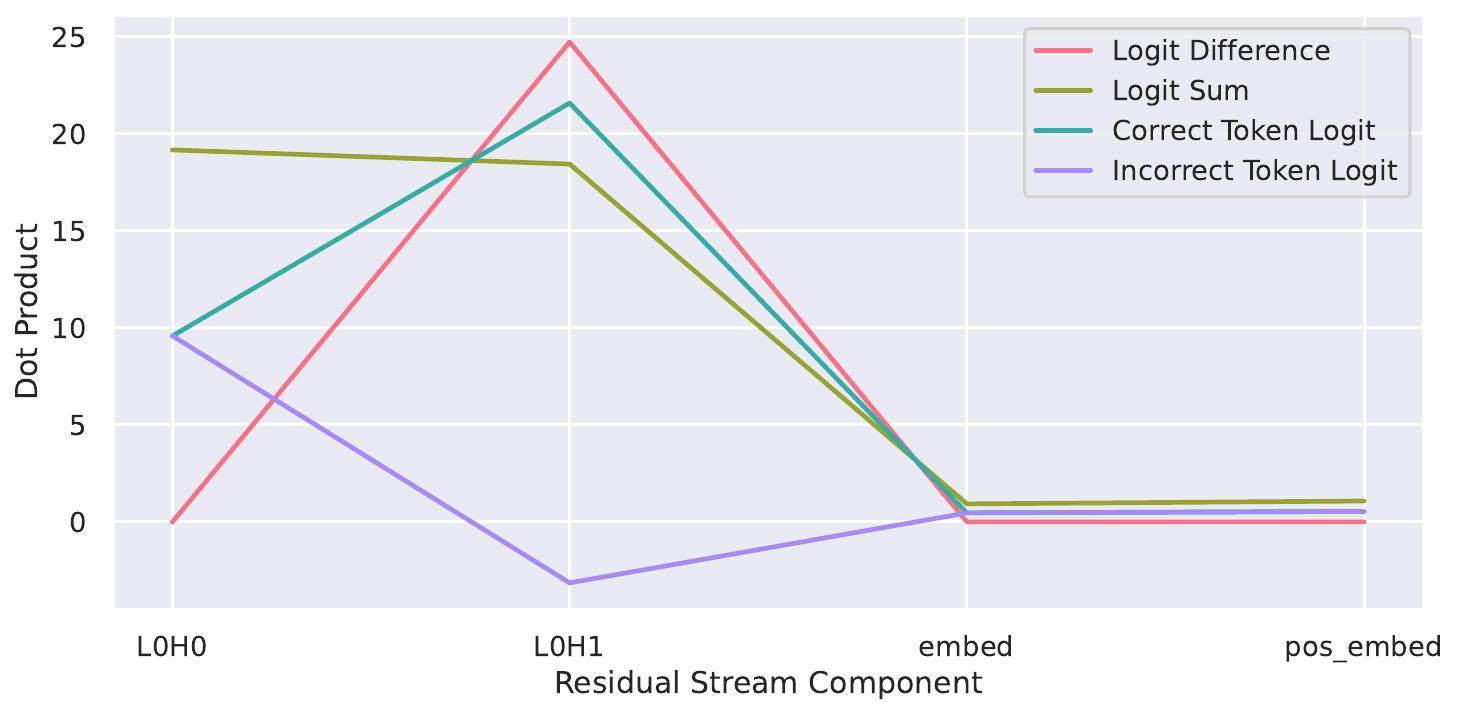}
    \caption{Residual stream decomposition for two-head, one-layer model. Projections of the residual stream components onto the correct and incorrect directions, along with their sum and difference.}
    \label{fig:residual_stream_decomposition_layers_1_heads_2}
\end{figure}

To understand how the model's components contribute to the final prediction, we decompose the residual stream at the final token position (corresponding to \texttt{<MID>} token) into the contributions of those components. We then project these components onto directions in the embedding space corresponding to the correct and incorrect names, as well as their sum and difference, using LogitLens~\citep{nostalgebraist2020interpreting}. \Cref{fig:residual_stream_decomposition_layers_1_heads_2} shows that the first head's output is aligned closest with the sum direction, i.e., it represents the combined contribution of both the correct and incorrect names (\textit{correct + incorrect}). On the other hand, the second head's output aligns closest with the direction of the token difference, i.e., the contrast between the two name embeddings (\textit{correct -- incorrect}). Since the final logits are computed by adding the contribution of all the components, the logit component for the incorrect token roughly cancels out, and the direction corresponding to the correct token is amplified.

This analysis is also not foolproof because we can see that the second head also has some component in the direction of the correct token, as well as the sum direction. Nevertheless, we observe a clear division of labor between the two heads: one aggregates signals (additive), while the other suppresses the incorrect alternative (contrastive). Together, they form an additive-contrastive circuit to produce a clean, interpretable mechanism for generating the correct logits.

\subsubsection{Spectral Analysis of QK and OV Circuits}
\label{sec:spectral_analysis}

\begin{figure}[h]
    \begin{subfigure}
        {\linewidth}
        \includegraphics[width=\linewidth]{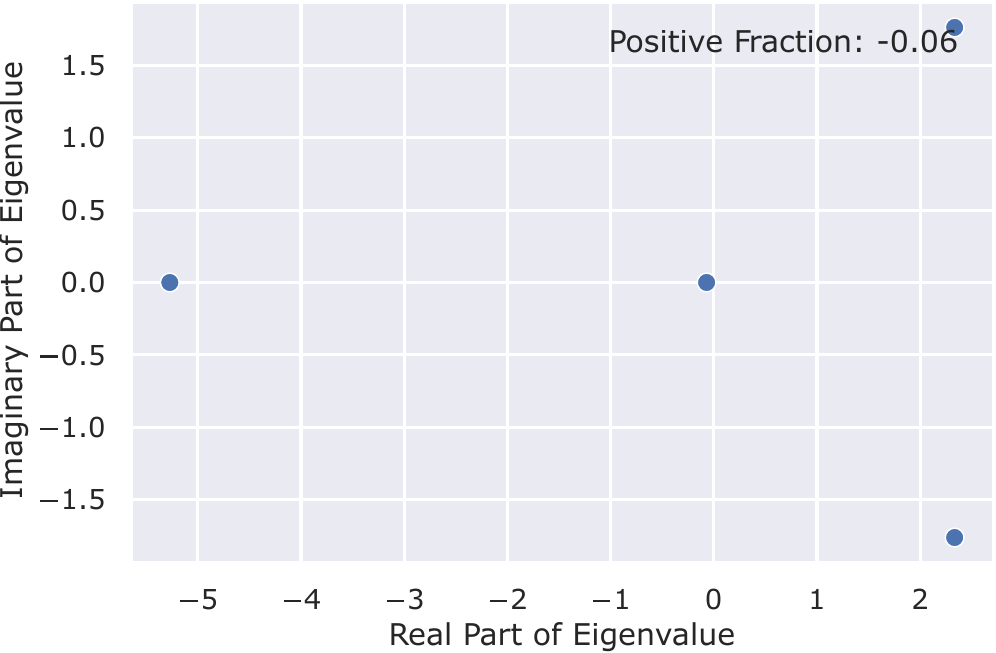}
        \caption{The first head exhibits relatively neutral dynamics.}
    \end{subfigure}
    \begin{subfigure}
        {\linewidth}
        \includegraphics[width=\linewidth]{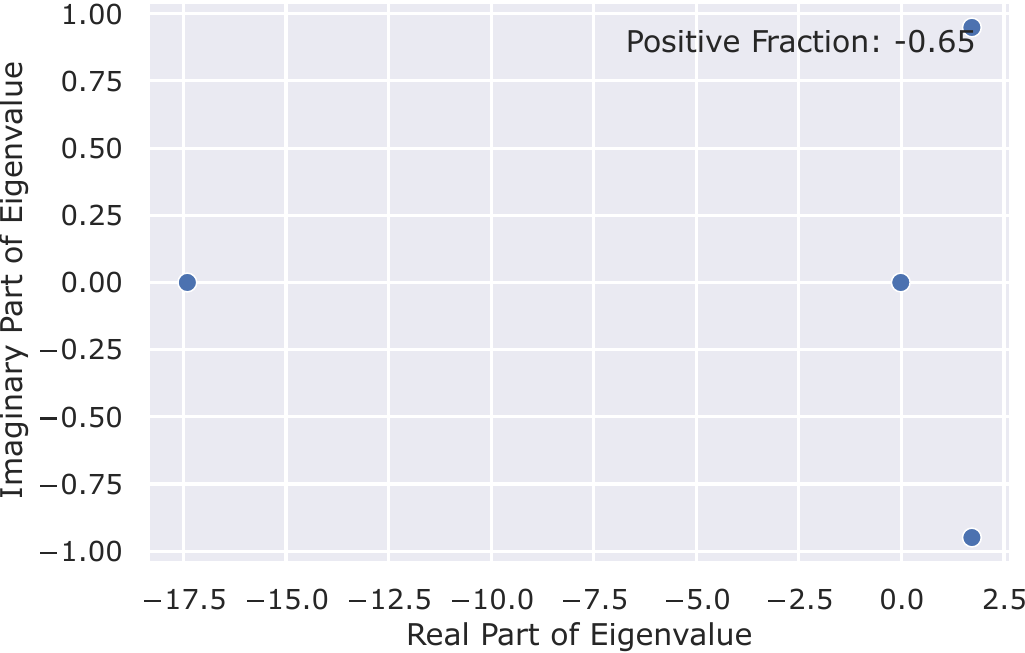}
        \caption{The second head indicates a strong suppressive effect.}
    \end{subfigure}
    \caption{Eigenvalue distribution of QK circuits for a two-heads, one-layer model.}
    \label{fig:qk_eigenvalues_layers_1_heads_2}
\end{figure}

\begin{figure}[h]
    \begin{subfigure}
        {\linewidth}
        \includegraphics[width=\linewidth]{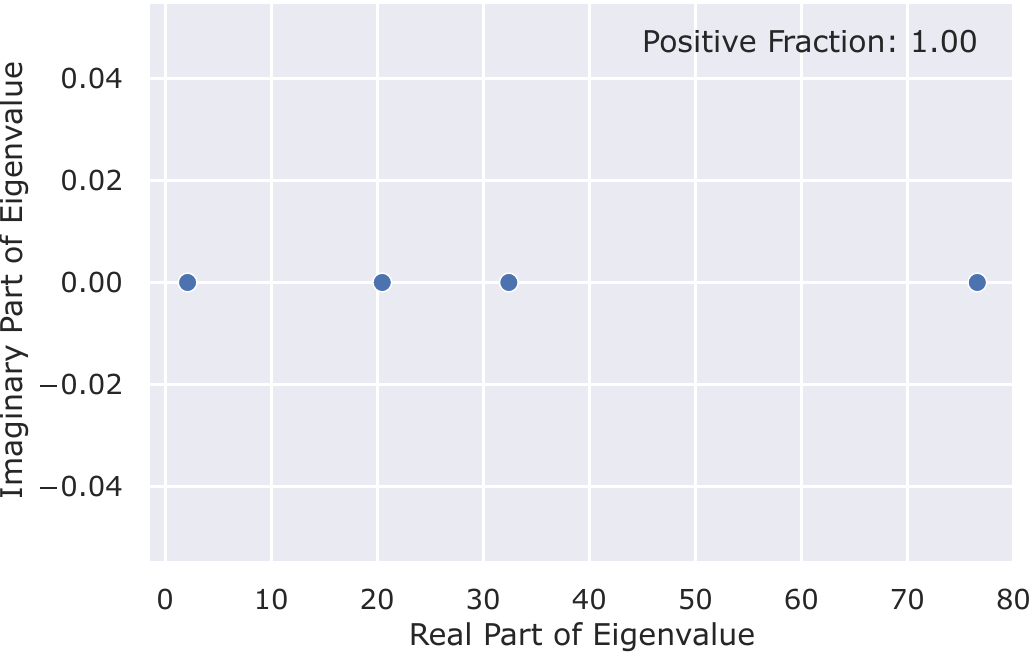}
        \caption{The first head acts purely additively.}
    \end{subfigure}
    \begin{subfigure}
        {\linewidth}
        \includegraphics[width=\linewidth]{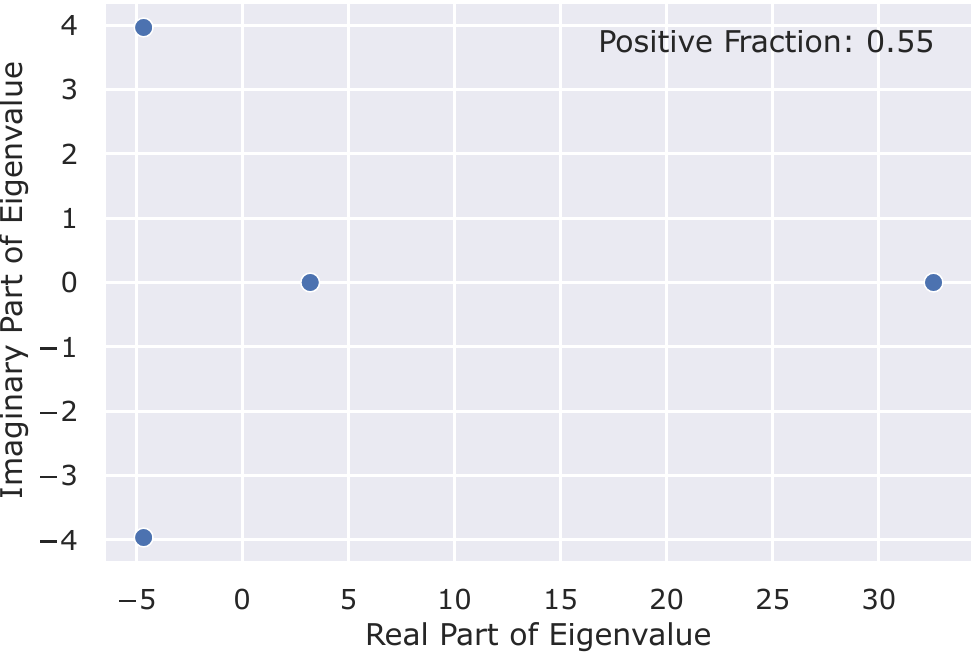}
        \caption{The second head suggests a mix of additive and contrastive (rotational) contributions.}
    \end{subfigure}
    \caption{Eigenvalue distribution of OV circuits for a two-heads, one-layer model.}
    \label{fig:ov_eigenvalues_layers_1_heads_2}
\end{figure}

While random matrices typically exhibit symmetric eigenvalue distributions \citep{tarnowski2022real}, the QK and OV matrices in our two-head model display significant asymmetry, reinforcing their specialized functional roles (see \cref{fig:ov_eigenvalues_layers_1_heads_2,fig:qk_eigenvalues_layers_1_heads_2}). Furthermore, on the top-right of each subfigures, we report the fraction of positive eigenvalues for each head calculated using the formula $\frac{\sum_{i}\lambda_{i}}{\sum_{i}\abs{\lambda_{i}}}$, where $\lambda_{i}$ are the eigenvalues of the matrix and $\abs{\lambda_{i}}$ are their magnitudes.

\paragraph{Spectral Properties of QK Circuits}

Observing the eigendecomposition of the QK matrices in \cref{fig:qk_eigenvalues_layers_1_heads_2}, we notice that the first head has a moderate suppression mechanism, denoted by a real eigenvalue of $\approx-5.2$, to forbid attending to some dimensions (or some tokens), indicating a less pronounced inhibitory effect. Additionally, it has two other complex eigenvalues $\approx 2.3 \pm 1.6\imath$, indicating some amplifying effect along with some rotation in some dimension. Finally, the positive fraction around zero ($-0.06$) suggests that the amplifying effect of the rotational component is almost balanced by the suppressive effect of the negative eigenvalue. Hence, the overall attention dynamics of the first head are relatively neutral.

The second head has a strong suppression mechanism ($\approx-17.5$), indicating a more pronounced inhibitory effect. The positive fraction of $-0.65$ suggests that the suppressive effect of the dominant negative eigenvalue outweighs the amplifying effect of the rotational components (denoted by complex eigenvalues), leading to an overall inhibitory attention dynamics.

% \begin{figure}[h]
%     \includegraphics[width=\linewidth]{qk_eigenvalues_layers_1_heads_2}
%     \caption{\textbf{Eigenvalue Distribution of QK Circuits for Two-Head, One-Layer Model.} The second head has a larger dominant negative eigenvalue (positive fraction of $-0.65$) compared to the first head (positive fraction of $-0.06$), indicating a stronger suppressive effect in the second head's attention dynamics.}
%     \label{fig:qk_eigenvalues_layers_1_heads_2}
% \end{figure}

\paragraph{Spectral Properties of OV Circuits}

The eigendecomposition of the OV matrices further reveals the asymmetry between the two heads (see \cref{fig:ov_eigenvalues_layers_1_heads_2}). The first head is a \textbf{copying} or \textbf{passthrough} head, which identifies important tokens via its QK circuit and then amplifies their presence in the residual stream without any rotation or inversion.

The second head has half of its eigenvalues as real and positive, while the other half are imaginary with negative real parts. This suggests that the chosen token will copy itself in some dimension and rotate with inversion in another dimension, which can be interpreted as subtracting from the logits of the other token with some added transformations.

This distinction corresponds naturally with the roles inferred from embedding projections: one head aggregates signals (additive), while the other suppresses the incorrect alternative (contrastive).

% \begin{figure}[h]
%     \includegraphics[width=\linewidth]{ov_eigenvalues_layers_1_heads_2}
%     \caption{\textbf{Eigenvalue Distribution of OV Circuits for Two-Head, One-Layer Model.} The first head has all positive eigenvalues (positive fraction of $1$), indicating an additive contribution to the residual stream. In contrast, the second head has positive as well as imaginary eigenvalues with negative real parts (positive fraction of $0.55$), suggesting a mix of additive and contrastive contributions.}
%     \label{fig:ov_eigenvalues_layers_1_heads_2}
% \end{figure}

\subsubsection{Positional Focus of Attention Heads}
\label{sec:positional_focus_layers_1_heads_2}

\begin{figure}[h]
    \begin{subfigure}
        {\linewidth}
        \includegraphics[width=\linewidth]{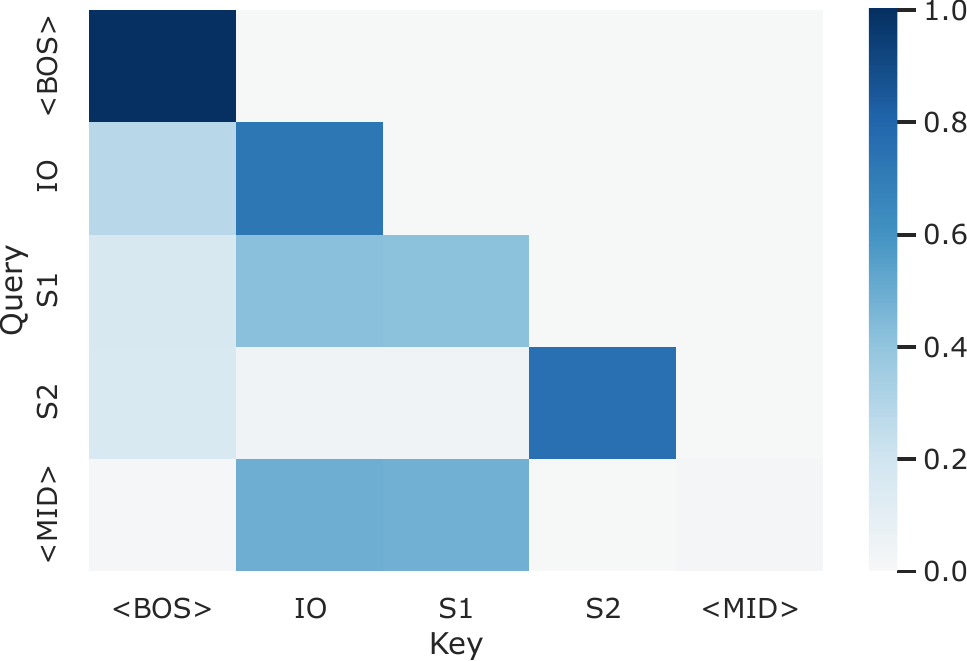}
        \caption{The first head focuses on the positional embeddings of the names in the dependent clause.}
        \label{fig:attention_patterns_position_layers_1_heads_2_head0}
    \end{subfigure}
    \begin{subfigure}
        {\linewidth}
        \includegraphics[width=\linewidth]{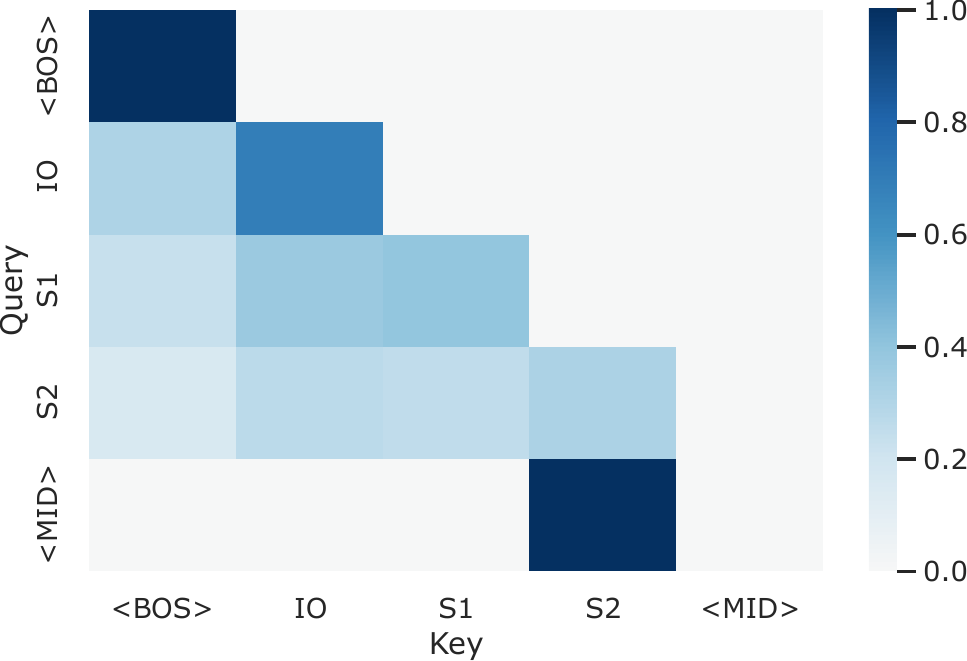}
        \caption{The second head attends primarily to the positional embedding of the subject of the main clause.}
    \end{subfigure}
    \caption{Average attention heatmap for two-head, one-layer model with name embeddings averaged.}
    \label{fig:attention_patterns_position_layers_1_heads_2}
\end{figure}

To isolate the model's reliance on positional embeddings, we assign a single, averaged embedding to all name tokens, removing the model's ability to differentiate them by identity. From \cref{fig:attention_patterns_position_layers_1_heads_2}, we can consider the first head as a \textbf{positional head} that focuses on the positions of the names in the dependent clause, independent of their word embeddings, because the attention patterns in \cref{fig:attention_patterns_ioi_layers_1_heads_2_head0,fig:attention_patterns_position_layers_1_heads_2_head0} look the same.

The second head attends predominantly to the position of occurrence of the subject in the main clause. However, despite this positional focus, it attends to the name in the dependent cause, not repeated in the main clause. This indicates that the second head is responsible for integrating positional as well as contextual information to determine the correct output.

% \begin{figure}[h]
%     \centering
%     \includegraphics[width=\linewidth]{attention_patterns_position_layers_1_heads_2}
%     \caption{\textbf{Average Attention Heatmap for Two-Head, One-Layer Model with Averaged Name Embeddings.} The first head focuses almost equally on the positional embeddings of the two name tokens from the dependent clause, while the second head attends primarily to the positional embedding of the subject token of the main clause.}
%     \label{fig:attention_patterns_position_layers_1_heads_2}
% \end{figure}

\subsubsection{Ablation: Positional Embeddings}

\begin{figure}[h]
    \begin{subfigure}
        {\linewidth}
        \includegraphics[width=\linewidth]{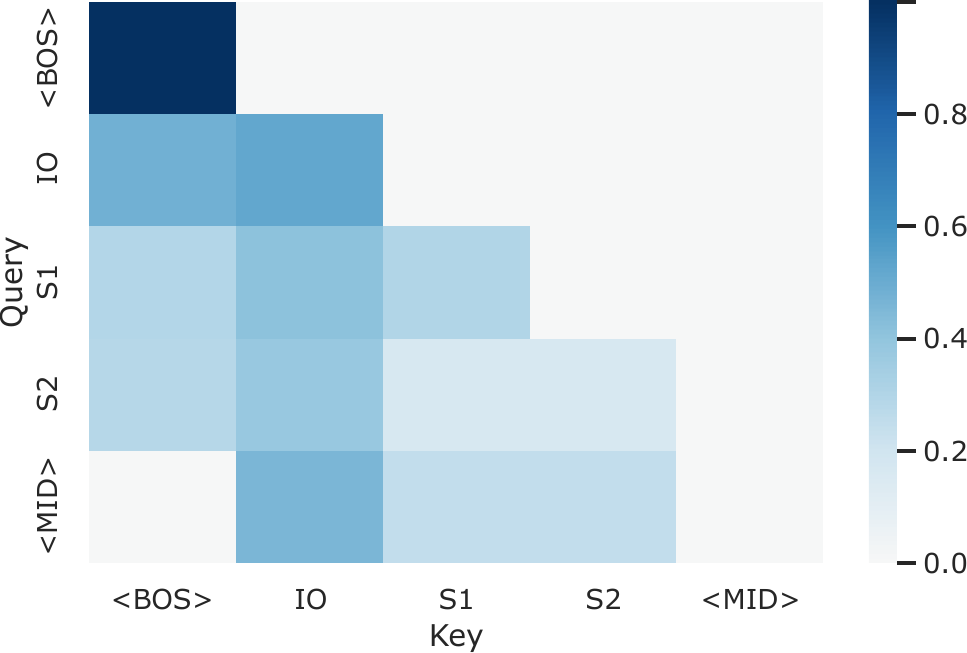}
        \caption{The head focuses only on the name tokens and most on IO for the ``BAAB'' template.}
    \end{subfigure}
    \begin{subfigure}
        {\linewidth}
        \includegraphics[width=\linewidth]{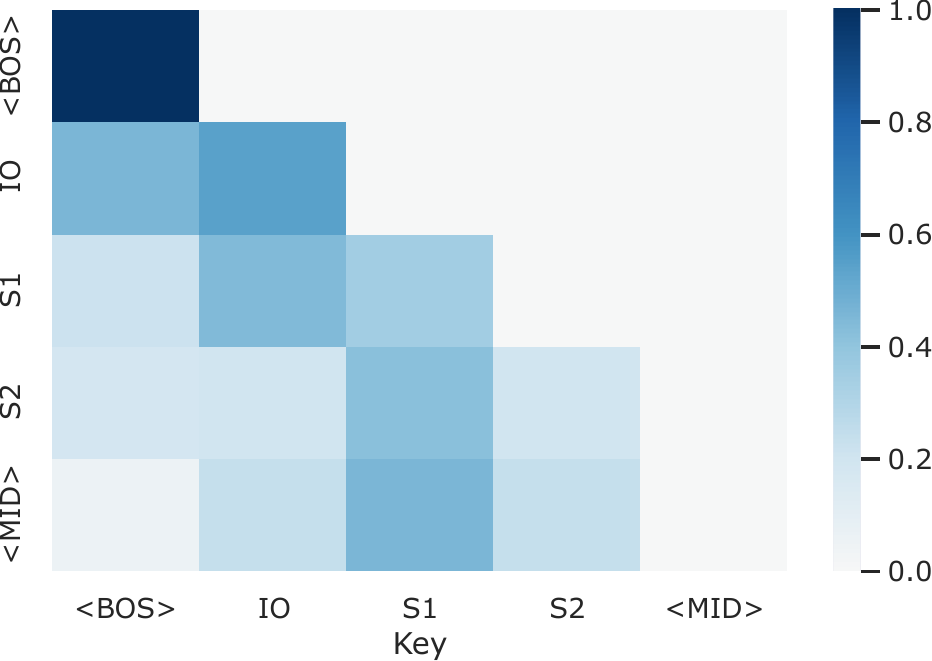}
        \caption{The head focuses only on the name tokens and most on S1 for the ``BABA'' template.}
    \end{subfigure}
    \caption{Average attention heatmap for the first head in a two-head, one-layer model trained without positional embeddings. Both heads focus on the name tokens of the prompt and focus most on the correct output name present in the dependent clause.}
    \label{fig:attention_patterns_ioi_layers_1_heads_2_no_pos}
\end{figure}

To study how the model utilizes positional information, we train the same model architecture without any positional embeddings. The model achieves an accuracy of $\approx70\%$ on the IOI task, with $\approx67\%$ probability on the correct token, indicating that positional embeddings are not strictly necessary for the model to learn the task. Unlike the ones with positional embeddings, the heads trained in this manner exhibit similar attention patterns, managing to focus primarily on the correct non-repeated name (see \cref{fig:attention_patterns_ioi_layers_1_heads_2_no_pos}). This suggests the model can fall back on purely semantic contextual relationships, though explicit positional cues drastically simplify the optimization landscape, allowing it to reach $100\%$ accuracy and perfectly decouple into additive-contrastive roles.

% \begin{figure}[h]
%     \centering
%     \includegraphics[width=\linewidth]{attention_patterns_ioi_layers_1_heads_2_no_pos}
%     \caption{\textbf{Average Attention Heatmap for Two-Head, One-Layer Model trained without Positional Embeddings.} Both heads focus mostly on the correct output name present in the dependent clause, i.e., \textit{B} in the ``BAAB'' template and \textit{A} in the ``BABA'' template.}
%     \label{fig:attention_patterns_ioi_layers_1_heads_2_no_pos}
% \end{figure}

\subsection{A Two-Layers, One-Head Model}

We also train a two-layer attention-only transformer with one head in each layer to observe how a model performs IOI in the availability of compositions~\citep{elhage2021mathematical}. The head dimension was $4$ for the one-layer two-heads model. For this model, since we have a single head per layer, the head dimension is the same as the hidden dimension, i.e., $8$. So, this model has more representational capacity than the one-head one-layer model. If this model doesn't perform any composition, i.e., if the second layer just doesn't depend on the output of the first layer, then it is the same as the one-layer two-heads model with each layer writing to the residual stream in orthogonal subspaces of $4$ dimensions each.

\subsubsection{Attention Heatmap}

% \begin{figure}[h]
%     \includegraphics[width=\linewidth]{attention_patterns_ioi_layers_2_heads_1_combined}
%     \caption{Attention heatmap for two-layer, one-head model. The heatmap of both heads changes depending on the template.}
%     \label{fig:attention_patterns_ioi_layers_2_heads_1}
% \end{figure}

\begin{figure}[t]
    \begin{subfigure}
        {\linewidth}
        \includegraphics[width=\linewidth]{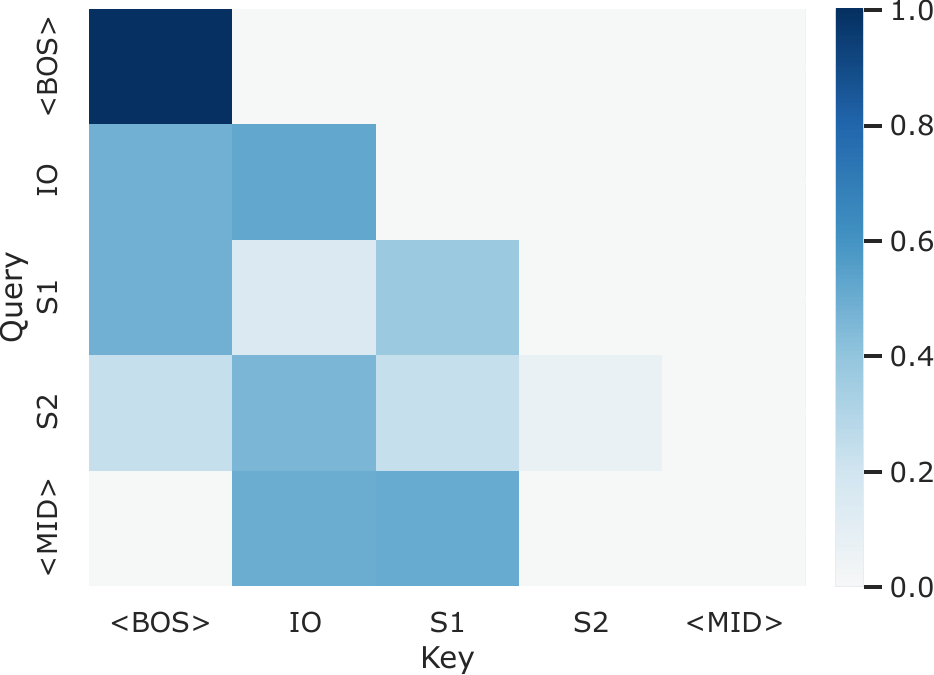}
        \caption{Attention heatmap for the first layer's head.}
        \label{fig:attention_patterns_ioi_layers_2_heads_1_baab_l0h0}
    \end{subfigure}
    \begin{subfigure}
        {\linewidth}
        \includegraphics[width=\linewidth]{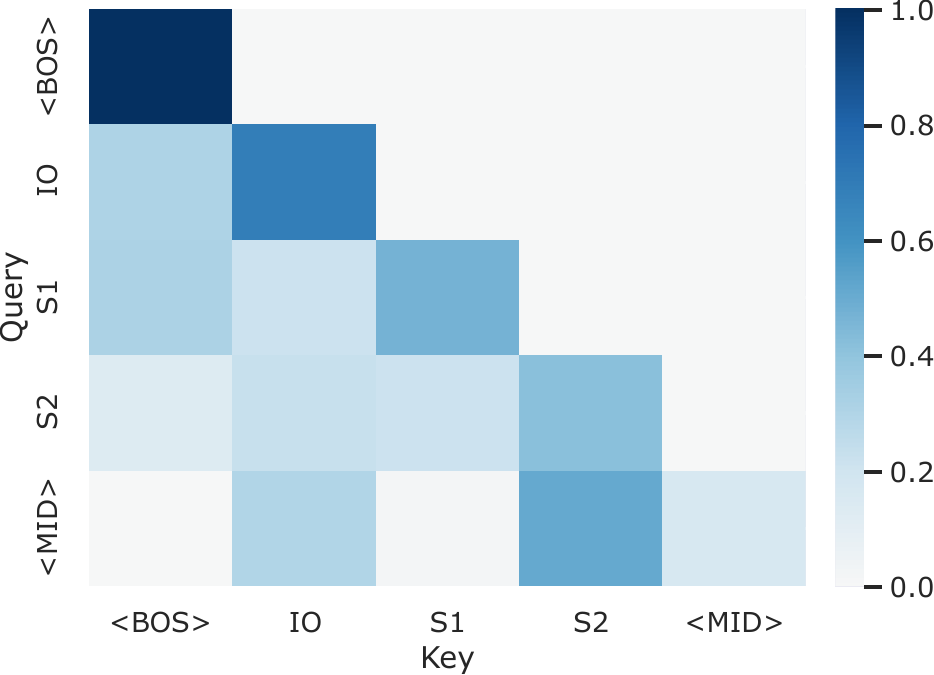}
        \caption{Attention heatmap for the second layer's head.}
        \label{fig:attention_patterns_ioi_layers_2_heads_1_baab_l1h0}
    \end{subfigure}
    \caption{Attention heatmap for a two-layer, one-head model for the ``BAAB'' template.}
    \label{fig:attention_patterns_ioi_layers_2_heads_1_baab}
\end{figure}

\begin{figure}[t]
    \begin{subfigure}
        {\linewidth}
        \includegraphics[width=\linewidth]{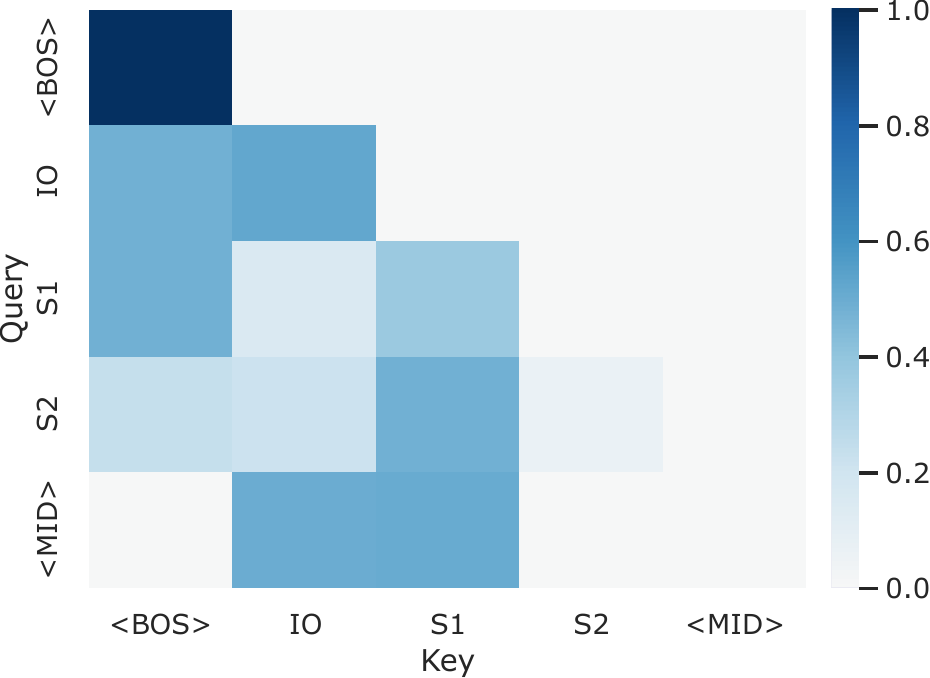}
        \caption{Attention heatmap for the first layer's head.}
        \label{fig:attention_patterns_ioi_layers_2_heads_1_baba_l0h0}
    \end{subfigure}
    \begin{subfigure}
        {\linewidth}
        \includegraphics[width=\linewidth]{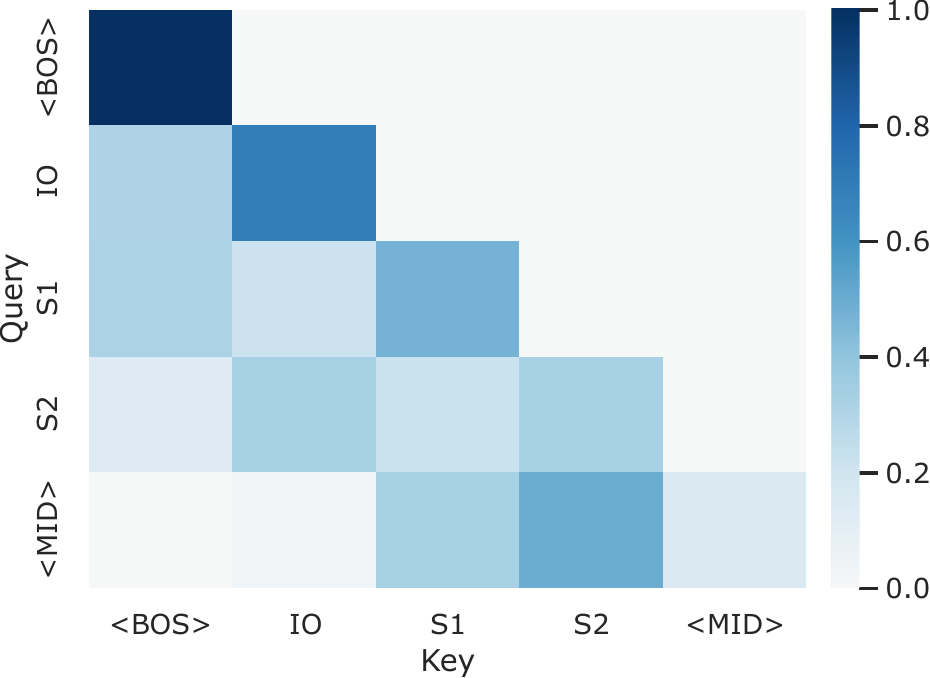}
        \caption{Attention heatmap for the second layer's head.}
        \label{fig:attention_patterns_ioi_layers_2_heads_1_baba_l1h0}
    \end{subfigure}
    \caption{Attention heatmap for a two-layer, one-head model for the ``BABA'' template.}
    \label{fig:attention_patterns_ioi_layers_2_heads_1_baba}
\end{figure}

% \begin{figure}[h]
%     \begin{subfigure}
%         {\linewidth}
%         \includegraphics[width=\linewidth]{attention_patterns_ioi_layers_2_heads_1_avg_l0h0}
%         \caption{Attention heatmap for the first layer's head.}
%         \label{fig:attention_patterns_ioi_layers_2_heads_1_avg_l0h0}
%     \end{subfigure}
%     \begin{subfigure}
%         {\linewidth}
%         \includegraphics[width=\linewidth]{attention_patterns_ioi_layers_2_heads_1_avg_l1h0}
%         \caption{Attention heatmap for the second layer's head.}
%     \end{subfigure}
%     \caption{Attention heatmap for a two-layer, one-head model, averaged across both templates.}
%     \label{fig:attention_patterns_ioi_layers_2_heads_1_avg}
% \end{figure}

\Cref{fig:attention_patterns_ioi_layers_2_heads_1_baab,fig:attention_patterns_ioi_layers_2_heads_1_baba} show the attention heatmap for a two-layer, one-head model, averaged across ``BAAB'' and ``BABA'' templates, respectively. We observe that the attention patterns of both layers change depending on the template.
% From \cref{fig:attention_patterns_ioi_layers_2_heads_1_baab,fig:attention_patterns_ioi_layers_2_heads_1_baba,fig:attention_patterns_ioi_layers_2_heads_1_avg}, we observe that the attention patterns of both layers change depending on the template.
From \cref{fig:attention_patterns_ioi_layers_2_heads_1_baab_l0h0,fig:attention_patterns_ioi_layers_2_heads_1_baba_l0h0}, we observe that the \texttt{<MID>} token in the first layer still attends to both the name tokens in the dependent clause for both templates, similar to the first head of the one-layer two-heads model. However, the \texttt{S2} token in the first layer changes its attention pattern based on the template; it attends more to the \texttt{IO} token than the \texttt{S1} token for the ``BAAB'' template and more to the \texttt{S1} than the \texttt{IO} token. So, the first head is not solely positional, but aggregates information to \texttt{S2} token to be used by the latter head. Although the attention pattern of the second layer for the \texttt{<MID>} token (see \cref{fig:attention_patterns_ioi_layers_2_heads_1_baab_l1h0,fig:attention_patterns_ioi_layers_2_heads_1_baba_l1h0}) seems almost similar to the attention pattern of the second head of the one-layer two-heads model (see \cref{fig:attention_patterns_ioi_layers_1_heads_2_combined}), this time it attends to the aggregated information from the first layer.

\subsubsection{Role of Positional Embeddings}

\begin{figure}[h]
    \begin{subfigure}
        {\linewidth}
        \includegraphics[width=\linewidth]{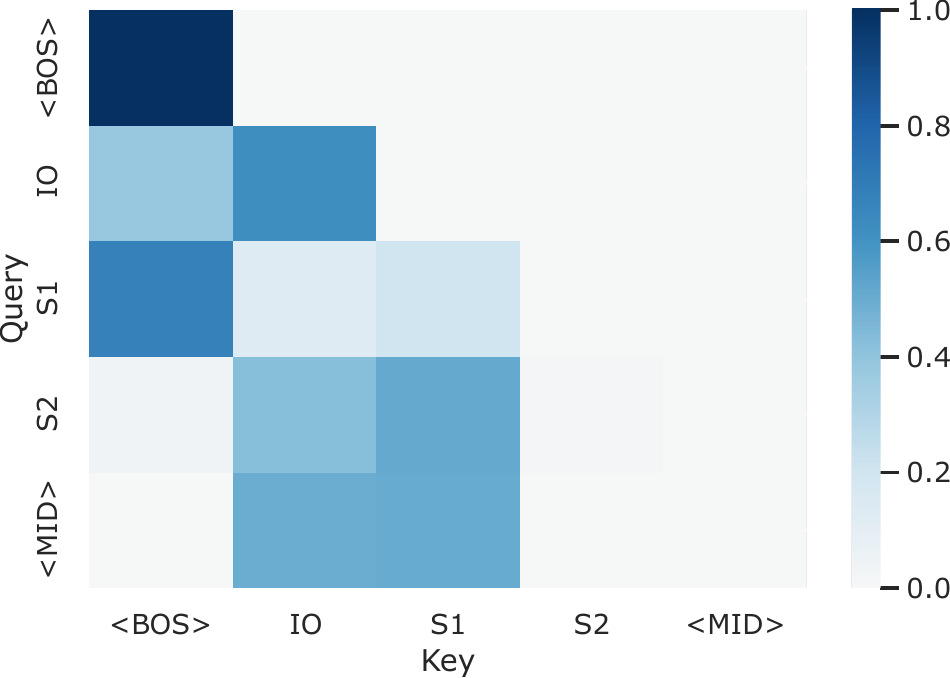}
        \caption{The head on the first layer is a positional head with almost the same attention pattern with indentifiable name embeddings.}
        \label{fig:attention_patterns_position_layers_2_heads_1_layer0}
    \end{subfigure}
    \begin{subfigure}
        {\linewidth}
        \includegraphics[width=\linewidth]{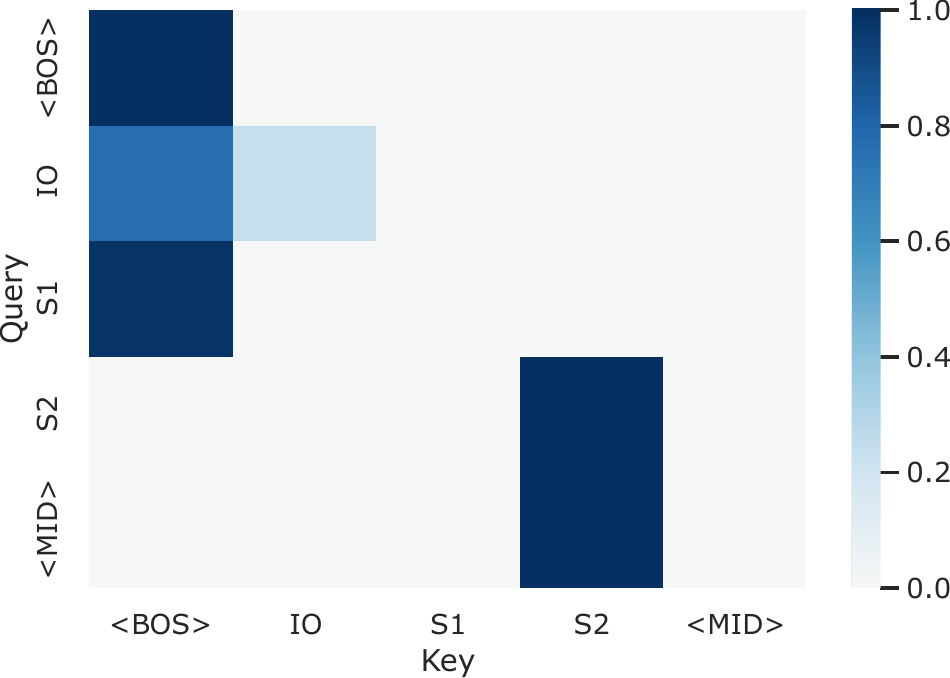}
        \caption{The head on the second layer pays strong attention to the positional embedding of the subject of the main clause.}
        \label{fig:attention_patterns_position_layers_2_heads_1_layer1}
    \end{subfigure}
    \caption{Attention heatmap for two-layer, one-head model with averaged name embeddings. We see a strong positional focus in both heads.}
    \label{fig:attention_patterns_position_layers_2_heads_1_horizontal}
\end{figure}

Similar to \cref{sec:positional_focus_layers_1_heads_2}, we analyze the attention pattern of the heads when the distinction among the names is removed (see \cref{fig:attention_patterns_position_layers_2_heads_1_horizontal}). We observe that the attention pattern in \cref{fig:attention_patterns_position_layers_2_heads_1_layer0} is very similar to \cref{fig:attention_patterns_position_layers_1_heads_2_head0}, indicating that it has a strong positional focus. However, although the \texttt{<MID>} token in the second layer (see \cref{fig:attention_patterns_position_layers_2_heads_1_layer1}) attends primarily to the token before it, it changes its attention pattern based on the context provided by the first layer. So, the second head is not solely positional, but integrates positional as well as contextual information to determine the correct output.

% \begin{figure}[h]
%     \centering
%     \includegraphics[width=\linewidth]{attention_patterns_position_layers_2_heads_1_horizontal}
%     \caption{\textbf{Attention Heatmap for Two-Layer, One-Head Model with Averaged Name Embeddings.} We see a strong positional focus in both heads.}
%     \label{fig:attention_patterns_position_layers_2_heads_1_horizontal}
% \end{figure}

\subsubsection{Ablation: Q, K, and V-Composition}
\label{sec:composition_ablation} To study the type of composition that the model is performing (Q, K, or V), we ablate them one by one by subtracting the output of the first layer from the corresponding input of the Q, K, and V matrices. We observe a drop in accuracy in the following order: Q-composition ($\approx100\%$ drop), V-composition ($\approx93.33\%$ drop), and K-composition ($\approx26.67\%$ drop). This indicates that the model is heavily relying on the Q and V-compositions to perform the task. So, we can conclude that the two-layer one-head model is indeed performing some composition to solve the IOI task, different from the one-layer two-heads model. This hints that finding a circuit capable of solving a given task using composition is an easier task for the optimizer than building two orthogonal subspaces in the residual stream.
    \section{Conclusion}

In this study, we showed that a single-head single-layer attention-only transformer can't solve a symbolic version of the Indirect Object Identification (IOI) task. However, if we increase the number of attention heads to two, keeping the number of parameters the same, it can perfectly solve it. Our mechanistic analysis revealed an elegant division of labor: one head aggregates referential information additively, while the other performs contrastive suppression of incorrect alternatives. In a two-layer, single-head model, we further observed compositional behavior across layers, indicating the emergence of functional hierarchy. These findings highlight that task-constrained training can produce parsimonious and interpretable circuits, offering valuable insight into the primitive computational motifs that may underlie reasoning in larger, pretrained language models.
    \section*{Limitations}

While this work successfully isolates minimal computational motifs for coreference-like reasoning, our analysis is bounded by the following constraints.

\paragraph*{Sensitivity to Sequence Structure}
By abstracting away linguistic complexity into rigid $6$-token sequences, we successfully isolated the core exclusionary logic of IOI. However, this paper doesn't explore how this minimal circuit behaves when subjected to varying sequence lengths, multiple interdependent clauses, or dynamic syntax, and at what threshold of structural complexity this two-head circuit necessitates the multi-hop mechanisms described by \citet{DBLP:conf/iclr/WangVCSS23}.

\paragraph*{Training Dynamics}
Our mechanistic analysis focuses exclusively on the fully converged model. We do not investigate the developmental interpretability or training dynamics that lead to the emergence of these specialized circuits. Specifically, it is currently unknown at what phase during the optimization process the two heads differentiate into their respective additive and contrastive roles, or what specific loss landscape dynamics drive this strict division of labor.
    % \input{latex/acknowledgments}

    % Include anthology-2, if you need references from before 2012
    \bibliography{references,anthology_subset}

    \appendix
\end{document}